\documentclass[sigconf,natbib=true]{acmart}
\usepackage{balance}
\usepackage{algorithm2e}
\usepackage{booktabs}
\usepackage{pbox, ragged2e}
\usepackage{adjustbox}
\usepackage{xcolor}
\AtBeginDocument{%
  }

\setcopyright{acmlicensed}
\copyrightyear{2026}
\acmYear{2026}
\acmConference[CHIIR '26]{Proceedings of the 2025 Conference on Human Information Interaction and Retrieval}{March 22--26, 2026}{Seattle, United States}

\begin{document}


\title[HARNESS: Human-Agent Risk Navigation and Event Safety System]{HARNESS: Human-Agent Risk Navigation and Event Safety System for Proactive Hazard Forecasting in High-Risk DOE Environments}


\author{Sanjay Das}
\authornote{Both authors contributed equally to this research.}
\affiliation{%
  \institution{Oak Ridge National Laboratory}
  \city{Oak Ridge}
  \state{Tennessee}
  \country{USA}}
\email{dass3@ornl.gov}

\author{Ran Elgedawy}
\authornotemark[1]
\affiliation{%
  \institution{Oak Ridge National Laboratory}
  \city{Oak Ridge}
  \state{Tennessee}
  \country{USA}}
\email{elgedawyr@ornl.gov}

\author{Ethan Seefried}
\affiliation{%
  \institution{Oak Ridge National Laboratory}
  \city{Oak Ridge}
  \state{Tennessee}
  \country{USA}}
\email{seefriedej@ornl.gov}

\author{Gavin Wiggins}
\affiliation{%
  \institution{Oak Ridge National Laboratory}
  \city{Oak Ridge}
  \state{Tennessee}
  \country{USA}}
\email{wigginsg@ornl.gov}

\author{Ryan Burchfield}
\affiliation{%
  \institution{Oak Ridge National Laboratory}
  \city{Oak Ridge}
  \state{Tennessee}
  \country{USA}}
\email{burchfieldra@ornl.gov}

\author{Dana Hewit}
\affiliation{%
  \institution{Oak Ridge National Laboratory}
  \city{Oak Ridge}
  \state{Tennessee}
  \country{USA}}
\email{hewitdm@ornl.gov}

\author{Sudarshan Srinivasan}
\affiliation{%
  \institution{Oak Ridge National Laboratory}
  \city{Oak Ridge}
  \state{Tennessee}
  \country{USA}}
\email{srinivasans@ornl.gov}

\author{Todd Thomas}
\affiliation{%
  \institution{Oak Ridge National Laboratory}
  \city{Oak Ridge}
  \state{Tennessee}
  \country{USA}}
\email{thomastm@ornl.gov}

\author{Prasanna Balaprakash}
\affiliation{%
  \institution{Oak Ridge National Laboratory}
  \city{Oak Ridge}
  \state{Tennessee}
  \country{USA}}
\email{pbalapra@ornl.gov}

\author{Tirthankar Ghosal}
\affiliation{%
  \institution{Oak Ridge National Laboratory}
  \city{Oak Ridge}
  \state{Tennessee}
  \country{USA}}
\email{ghosalt@ornl.gov}



\renewcommand{\shortauthors}{Das and Elgedawy et al.}

\begin{abstract}
Operational safety at mission-critical work sites is a top priority given the complex and hazardous nature of daily tasks. This paper presents the Human-Agent Risk Navigation and Event Safety System (HARNESS), a modular AI framework designed to forecast hazardous events and analyze operational risks in U.S. Department of Energy (DOE) environments. HARNESS integrates Large Language Models (LLMs) with structured work data, historical event retrieval, and risk analysis to proactively identify potential hazards. A human-in-the-loop mechanism allows subject matter experts (SMEs) to refine predictions, creating an adaptive learning loop that enhances performance over time. By combining SME collaboration with iterative agentic reasoning, HARNESS improves the reliability and efficiency of predictive safety systems. Preliminary deployment shows promising results, with future work focusing on quantitative evaluation of accuracy, SME agreement, and decision latency reduction.
\end{abstract}




\maketitle
\authorsaddresses{
\textbf{Affiliation for all authors:} Oak Ridge National Laboratory, Oak Ridge, Tennessee, USA.
}
\section{Introduction}\label{sec:intro}

Operational safety at high-risk sites demands proactive and rigorous hazard identification. Traditional assessment methods are often time-consuming and prone to missing emerging risks. Advances in Artificial Intelligence (AI), particularly Large Language Models (LLMs), enable a paradigm shift by rapidly analyzing vast, unstructured data. Their natural language understanding makes them well-suited for extracting insights from complex incident reports and risk narratives.

Our system introduces a novel approach to predictive hazard identification by integrating advanced information retrieval with LLM-driven risk analysis. This fusion provides early foresight into potential hazards, supporting safer and more informed work planning. The goal is to transform safety management from a reactive process into a predictive framework that anticipates and mitigates risks before incidents occur.

Prior work has leveraged ML for incident frequency analysis \cite{Mojaddadi15122017, PALTRINIERI2019475, TAMASCELLI2022107786} and LLMs for textual safety data processing \cite{BAEK2025106255, 10.1145/3674805.3695401, unknownMumtarin, app14041352}. However, traditional single-prompt LLM solutions struggle with traceability, regulatory drift, and contextual specificity \cite{ray2025survey}, leading to issues like hallucination or a lack of focus on mission-critical details. We address this by proposing a multi-agent orchestration that unbundles retrieval, reasoning, validation, and human feedback into discrete, cooperative agents. This structure enables rigorous coordination, enhances traceability, and provides a clear mechanism for human-in-the-loop refinement, going beyond standard RAG systems to deliver a trustworthy, auditable, and adaptive system for high-consequence environments.

Our contributions are as follows:
\begin{itemize}
    \item We present the HARNESS system, a novel, multi-agent retrieval augmented generation (RAG) pipeline for proactive hazard forecasting in high-risk operational environments.

    \item We propose a smart RAG system for intelligent retrieval of relevant incidents data from a vector database.

    \item We integrate Standards-Based Management System (SBMS) and external control policies for providing appropriate mitigation approaches for identified critical hazards.

    \item HARNESS generates comprehensive vulnerability reports combining all the relevant past events, critical missing hazards and controls towards presenting an overall risk profile of the corresponding work plan.
    
\end{itemize}

\section{Dataset}\label{sec:dataset}

\subsection{Dataset Overview}
Our corpus combines four safety and incident reporting datasets from multiple DOE institutions, totaling \textbf{65,107} documents spanning over three decades (1990–2024) of safety-related reporting. These sources include structured reports, narrative-style lessons-learned records, occupational injury and illness reports, and localized site-specific reports.

\subsection{Target Users}
The primary target users are \textbf{Work Planners}, and \textbf{Subject Matter Experts (SMEs)} at high-consequence US Department of Energy (DOE) environments national laboratories. These users require accurate, fast, and grounded analysis of work plans against historical hazards and regulatory controls to ensure mission safety.

\subsection{Dataset Statistics}
Each document in the dataset includes metadata such as event name, date, location, a text summary, and a full body text, among other attributes. The summaries had an average length of $231.63 \pm 361.36$ words, with a median of 73 words and a maximum of 3,476 words. The full body texts averaged $1,677.22 \pm 3,246.98$ words, with a median of 678 words and a maximum of 63,337 words.


\section{System Architecture}\label{sec:method}
\subsection{Agentic RAG Workflow}
We propose a Retrieval-Augmented Generation (RAG)~\cite{lewis2020retrieval} system that integrates high-dimensional embedding-based search, multi-source semantic augmentation, and an intelligent query orchestration mechanism built upon GPT-4o~\cite{islam2025gpt} and complementary tools such as Perplexity~\cite{gravina2024charting}. 

We designed a RAG system empowered by a Smart RAG Agent, which orchestrates the retrieval pipeline using embedding-based search and GPT-4o-driven reasoning. The system accepts heterogeneous user queries $Q$, which may be plain text $(Q_{\text{text}})$, a single document $(Q_{\text{doc}})$, or a set of documents $(Q_{\text{docs}} = \{q_1, q_2, \ldots, q_k\})$. The corpus $\mathcal{D} = \{d_1, d_2, \ldots, d_N\}$ is segmented into chunks $C = \{c_1, c_2, \ldots, c_M\}$, with each chunk embedded using embedding size of 1024 tokens in $\mathbb{R}^d$ using a function $f_{\text{embed}} : \text{Text} \rightarrow \mathbb{R}^d$.

Upon receiving the query, the Smart RAG Agent interprets it semantically using GPT-4o and auxiliary tools (e.g., Perplexity), yielding an intent vector $v_Q = f_{\text{sem}}(Q)$. If the query is compound or underspecified, it is decomposed into subqueries $Q = \{Q_1, Q_2, \ldots,  Q_m\}$. Each subquery may be expanded into multiple formulations via semantic entailment and paraphrasing:

\begin{equation}
Q_i^{\text{exp}} = \text{Expand}(Q_i) = \{q_i^1, q_i^2, \ldots, q_i^k\}
\end{equation}

while ensuring semantic closeness:

\begin{equation}
\forall q_i^j, \quad \text{sim}(f_{\text{embed}}(q_i^j), f_{\text{embed}}(Q_i)) \geq \tau
\end{equation}

where $\tau$ is typically set to 0.8 cosine similarity.

Each subquery $q_i^j$ is embedded and used to search a vector database for the top-$K$ relevant chunks:

\begin{equation}
R_i^j = \text{top-$K$}_{c \in C} \left( \cos\left(f_{\text{embed}}(q_i^j), f_{\text{embed}}(c)\right) \right)
\end{equation}

All results are aggregated into $R = \bigcup_{i, j} R_i^j$ and filtered to remove low-similarity candidates (candidates below a threshold 'theta' of 0.5):

\begin{equation}
R_{\text{filtered}} = \{c \in R \mid \max_{q \in Q} \cos(f_{\text{embed}}(c), f_{\text{embed}}(q)) \geq \theta\}
\end{equation}

To ensure semantic precision, each remaining chunk is reranked using a cross-encoder:

\begin{equation}
s_i = f_{\text{cross}}(Q, c_i), \quad \forall c_i \in R_{\text{filtered}}
\end{equation}

and sorted by score to form the candidate list:

\begin{equation}
R_{\text{final}} = \text{Sort}_{c_i \in R_{\text{filtered}}}(s_i)
\end{equation}

We select the top-$k$ final chunks.
The selected context $\mathcal{C}_{\text{out}} = \{c_1^{\text{final}}, c_2^{\text{final}}, \ldots, c_k^{\text{final}}\}$ is injected into GPT-4o's input, enabling context-aware generation:

\begin{equation}
\text{Answer} = f_{\text{gen}}(Q, \mathcal{C}_{\text{out}})
\end{equation}

This architecture supports rich semantic reasoning, intelligent query reformulation, and optimized context construction.
\subsection{Agentic AI Risk Analysis Framework}
\begin{figure*}[!t]
    \centering
    \includegraphics[width=1\linewidth]{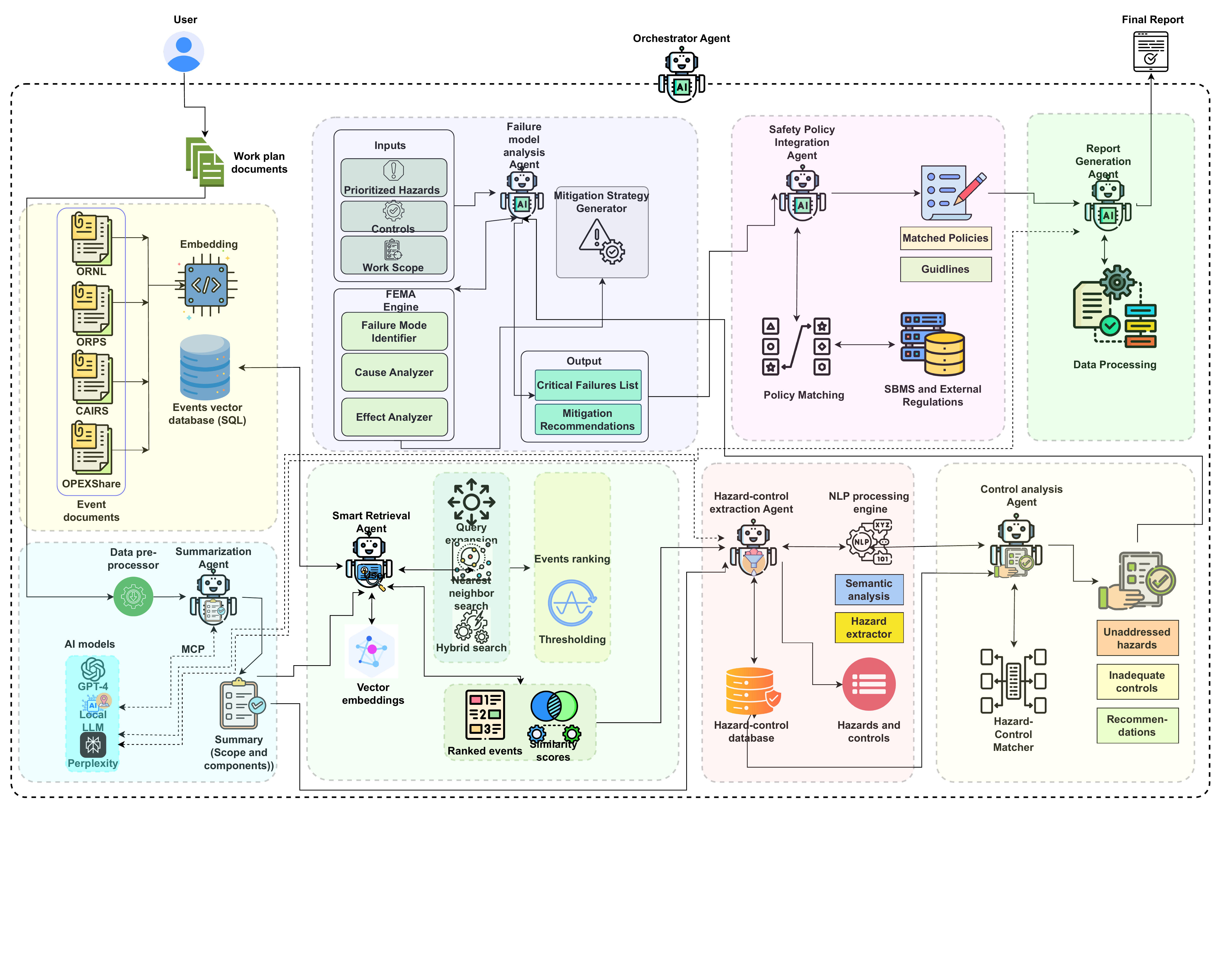}
    \Description[test]{test}
    \vspace{-26mm}
    \caption{HARNESS system architecture.}
    \label{fig:architecture}
    \vspace{-3mm}
\end{figure*}

Figure~\ref{fig:architecture} presents a modular, agentic architecture engineered for high-fidelity risk analysis, hazard identification, and mitigation strategy generation in safety-critical environments. The framework is orchestrated by a central \emph{Orchestrator Agent}, ensuring rigorous coordination across AI models. 

\subsubsection{Data Ingestion and Vectorization}
Unstructured event documentation is ingested from sources such as described in Section \ref{sec:dataset}. A pre-processing pipeline standardizes the textual content, which is then semantically embedded using transformer-based models. Resultant vector representations are persisted within a relational or approximate nearest-neighbor vector database to support efficient similarity retrieval.

\subsubsection{Summarization}

The \emph{Summarization Agent} is responsible for generating structured summaries from input documents, capturing critical information such as scope, work components, and operational context. This agent interfaces with advanced foundation models, including GPT-4o, and external tools like Perplexity, to enhance understanding of the text. 

\subsubsection{Smart Retrieval and Ranking}
The \emph{Smart Retrieval Agent} executes via a sequence of operations: semantic interpretation, query expansion, and nearest-neighbor indexing. Retrieved candidate event vectors are scored and thresholded by cosine similarity, producing a ranked set of relevant events reflecting semantic proximity and contextual relevance.

\subsubsection{Failure Mode Analysis Subsystem}
A dedicated \emph{Failure Model Analysis Agent}  integrates with a domain-specific Failure-Effects Mode Analysis (FEMA) engine comprising:
\begin{itemize}
  \item \text{Failure Mode Identifier}
  \item \text{Cause Analyzer}
  \item \text{Effect Analyzer}
\end{itemize}
This subsystem synthesizes a \emph{Critical Failures List}, corresponding \emph{Effects} and \emph{Mitigation Recommendations}, which are forwarded for further analysis and policy alignment.

\subsubsection{Hazard-Control Extraction and Assessment}
The \emph{Hazard- Control Extraction Agent}, utilizing an NLP engine, conducts semantic parsing to extract hazard-control pairs. These pairs are stored in a dedicated \emph{Hazard-Control Database}. Subsequently, the \emph{Control Analysis Agent} employs a Hazard-Control Matcher to analyze coverage. 

\subsubsection{Policy Matching and Mitigation Alignment}
The \emph{Safety Policy Integration Agent} aligns critical failures and mitigation proposals with SBMS policies and external regulatory requirements. A policy-matching engine ensures each recommendation is traceably mapped to relevant guidelines and standards.

\subsubsection{Report Generation and Output Consolidation}
The \emph{Report Generation Agent} aggregates all processed outputs—including identified hazards, failure modes, mitigations, and policy mappings—into a comprehensive \emph{Final Vulnerability Analysis Report}. Data formatting, structuring, and compliance validation are executed to ensure the document is suitable for operational stakeholders.
\section{Evaluation and Results}\label{sec:results}
This section details the choice of embedding models and the evaluation results of the different operations of the tool such as events retrieval and vulnerability report generation.

\subsection{Evaluation of Embedding Models}
To identify the optimal embedding model for HARNESS, we evaluated three high-performing embeddings based on their Hugging Face LLM leaderboard rankings \cite{huggingface_leaderboard2_2023}: SFR-Embedding-Mistral \cite{SFRAIResearch2024}, OpenAI text-embedding-3-large \cite{openai-embedding3large}, and INF-Retriever-v1 \cite{infly-ai_2025}.
\subsubsection{Experiment Setup}
We randomly selected 15 ORPS documents and generated 100 QA pairs through manual and AI-assisted annotation as reference sets. Performance metrics included:
\begin{enumerate}
\item \textbf{Answer Correctness:} Factual agreement with reference answers (RAGAS \cite{es-etal-2024-ragas})
\item \textbf{Average Query Time:} Mean latency including retrieval and generation
\end{enumerate}
\subsubsection{Results}
Table~\ref{tab:new_results} shows Qwen3-Embedding-8B achieved the best balance with 75.3\% correctness and 2.0s query time. SFR-Embedding-Mistral had 67.1\% correctness with the fastest time (1.9s), while INF-Retriever-v1 showed 68.1\% correctness but slowest performance (2.7s). OpenAI's model had lowest correctness (60.1\%) at 2.1s. Based on these results, we selected Qwen3-Embedding-8B for subsequent experiments.

\begin{table}[h!t]
\centering
\caption{Answer correctness and average query time for each embedding model.}
\resizebox{\columnwidth}{!}{%
\begin{tabular}{|l|c|c|}
\hline
\textbf{Model} & \textbf{Correctness (\%)} & \textbf{Avg Query Time (s)} \\ \hline \hline
Qwen3-Embedding-8B & 75.3 & 2.0 \\
SFR-Embedding-Mistral & 67.1 & 1.9 \\
INF-Retriever-v1 & 68.1 & 2.7 \\
OpenAI text-embedding-3-large & 60.1 & 2.1 \\ \hline
\end{tabular}
}
\vspace{-6mm}
\label{tab:new_results}
\end{table}

\subsection{Retrieval Performance Evaluation}
To assess retrieval accuracy without ground-truth datasets, we applied the pooled judgment method \cite{sparck1975report, voorhees2000variations, tonon2015pooling}, which builds relevance assessments by merging outputs from multiple retrieval variants \cite{sanderson2005information}. This approach, widely used in large-scale evaluations such as TREC \cite{harman1995overview,arguello2023overview}, is suitable for specialized domains where exhaustive labeling is infeasible.


We compared six retrieval variants: (1) \textit{current\_best} -- our hybrid system using LLM-generated keywords, document titles, and CrossEncoder reranking; (2) \textit{keywords\_only}; (3) \textit{pure\_rag} -- semantic similarity search; (4) \textit{title\_only}; (5) \textit{rule\_keywords} -- TF-IDF and NER-based extraction; and (6) \textit{extended\_keywords} -- hybrid with 10 keywords. For five test work plans, each variant retrieved the top 10 documents; up to 25 unique results per query were manually annotated on a three-point relevance scale (0--2).


\begin{table}[h!t]
\centering
\caption{Retrieval Performance Evaluation Results Across Six System Variants.}
\label{tab:retrieval_evaluation}
\begin{tabular}{lccccc}
\toprule
\textbf{System Variant}  & \textbf{P@5} & \textbf{R@5} & \textbf{F1@5}  \\
\midrule
RAG + keywords            & 0.920 $\pm$ 0.160 & 0.243 $\pm$ 0.053 & 0.384 $\pm$ 0.080  \\
Title only            & 0.880 $\pm$ 0.160 & 0.234 $\pm$ 0.058 & 0.369 $\pm$ 0.086  \\
Rule + keywords         & 0.800 $\pm$ 0.219 & 0.212 $\pm$ 0.068 & 0.334 $\pm$ 0.104  \\
Keywords only         & 0.760 $\pm$ 0.196 & 0.196 $\pm$ 0.042 & 0.311 $\pm$ 0.069  \\
Extended keywords      & 0.720 $\pm$ 0.271 & 0.184 $\pm$ 0.065 & 0.293 $\pm$ 0.104  \\
Pure RAG              & 0.680 $\pm$ 0.371 & 0.177 $\pm$ 0.101 & 0.281 $\pm$ 0.158  \\
\bottomrule
\end{tabular}
\end{table}

As shown in Table \ref{tab:retrieval_evaluation}, the hybrid system achieved the best results (F1@5 = 0.384 $\pm$ 0.080). The \textit{title\_only} variant performed closely (F1@5 = 0.369), highlighting the discriminative value of work plan titles.

\subsection{Generated Report Evaluation}

We assessed report quality using an LLM-as-Judge framework, following prior work showing that large language models can reliably evaluate text quality~\cite{zheng2024judging, dubois2024lengthcontrolled, kim2024prometheus}.

Twenty randomly selected reports were evaluated with GPT-4 as the judge model, each compared to its corresponding work plan. Evaluations covered five criteria: \emph{clarity} (use of technical terms), \emph{completeness} (coverage of hazards, lessons, and mitigations), \emph{usefulness} (support for decision-making), \emph{accuracy} (factual grounding), and \emph{specificity} (relevance to the work plan). Each criterion was rated on a 5-point Likert scale (1 = Poor, 5 = Excellent), with both numeric scores and textual justifications.

\begin{table}[h!t]
\centering
\caption{Mean LLM-as-Judge ratings for generated risk reports (Likert 1--5).}
\label{tab:llm_judge_results}
\begin{tabular}{lc}
\hline
\textbf{Dimension} & \textbf{Mean Rating} \\
\hline
Clarity & 4.0 \\
Completeness & 3.0 \\
Usefulness & 4.0 \\
Accuracy & 5.0 \\
Specificity & 3.0 \\
\hline
\textbf{Overall} & \textbf{3.8} \\
\hline
\end{tabular}
\end{table}

\emph{Accuracy} received a perfect score (5.0), with GPT-4 noting that reports ``consistently base hazard identifications on documented events and established safety protocols.'' This confirms that retrieval grounding effectively prevents unsupported claims.

\emph{Clarity} and \emph{usefulness} also scored high (4.0), reflecting precise terminology and actionable recommendations. However, lower scores for \emph{completeness} and \emph{specificity} (3.0 each) indicate that while major hazards were captured, some nuanced or facility-specific risks were missed. Future improvements should focus on deeper context extraction and more targeted hazard analysis.

\section{Demonstration and Conclusion}\label{sec:con}
The demonstration showcases an end-to-end HARNESS workflow where an SME submits a complex, high-risk work plan for automated risk assessment and hazard forecasting\footnote{Demo link: \url{https://colostate-my.sharepoint.com/:v:/g/personal/eseefrie_colostate_edu/EbwxMIm-bQ1ArOwY7CQ5ZxUBh0Wwxr1GKz0xxhZKgWtJuw?e=xRbFxQ}}. HARNESS generates interpretable risk profiles and vulnerability reports, enabling experts to identify overlooked hazards and refine mitigations.  

\textbf{Technical Impact:} HARNESS advances predictive safety in high-consequence domains by making LLM-based risk analysis \emph{auditable}, \emph{traceable}, and \emph{adaptive} through agentic orchestration and SME feedback. It provides real-time, explainable insights that improve reliability, cut decision latency, and strengthen trustworthy AI-driven safety in DOE operations.

\section*{Acknowledgements}
This research is sponsored by the Office of the Laboratory Director, Oak Ridge National Laboratory's Operational Excellence Initiatives, which is supported by the United
States Department of Energy (DOE)’s Office of Science under Contract No. DE-AC05-00OR22725.



\bibliographystyle{ACM-Reference-Format}
\bibliography{sample-base}


\end{document}